\documentclass{article}

\usepackage{graphicx} 
\usepackage{subfigure} 

\usepackage{natbib}

\usepackage{algorithm}
\usepackage{algorithmic}

\usepackage{hyperref}

\usepackage[accepted]{icml2013}

\usepackage{amsmath}
\usepackage{amsfonts}

\icmltitlerunning{Multi-Label Classifier Chains for Bird Sound}

\begin{document} 

\twocolumn[
\icmltitle{Multi-Label Classifier Chains for Bird Sound}

\icmlauthor{Forrest Briggs}{briggsf@eecs.oregonstate.edu}
\icmlauthor{Xiaoli Z. Fern}{xfern@eecs.oregonstate.edu}
\icmlauthor{Jed Irvine}{irvine@eecs.oregonstate.edu}
\icmladdress{Oregon State University, Corvallis, OR, 97333, USA}

\icmlkeywords{boring formatting information, machine learning, ICML}

\vskip 0.3in
]

\begin{abstract} 

Bird sound data collected with unattended microphones for automatic surveys, or mobile devices for citizen science, typically contain multiple simultaneously vocalizing birds of different species. However, few works have considered the multi-label structure in birdsong. We propose to use an ensemble of classifier chains combined with a histogram-of-segments representation for multi-label classification of birdsong. The proposed method is compared with binary relevance and three multi-instance multi-label  learning (MIML) algorithms from prior work (which focus more on structure in the sound, and less on structure in the label sets). Experiments are conducted on two real-world birdsong datasets, and show that the proposed method usually outperforms binary relevance (using the same features and base-classifier), and is better in some cases and worse in others compared to the MIML algorithms.

\end{abstract}

\section{Introduction}


The most familiar formulation of supervised classification associates single feature-vectors with single labels, hence it is called single-instance single-label (SISL). For example, SVM and logistic regression are SISL classifiers. One common setup involving SISL classifiers is to use a segmentation algorithm to extract ``syllables'' or calls of bird sound from a recording, each of which is described by a feature vector. A SISL classifier is trained on a collection of syllables paired with species labels, then predicts the species for a new syllable \cite{fagerlund2007bird,damoulas2010bayesian}.

Many of the audio recordings used in SISL experiments are collected with a directional microphone aimed by a person at the bird of interest. This method produces recordings where the targeted bird is louder than other sound sources in the environment. 
Audio data collected by unattended microphones for the purpose of acoustic monitoring, and audio collected with mobile devices for citizen science are less ideal; it is common to have multiple simultaneously vocalizing bird species, in addition to other sources of noise such as non-bird species, wind, rain, streams, and motor vehicles. Few works have addressed these complexities in real-world data \cite{brandes2008feature,briggs2012acoustic}.

There are two kinds of structure in bird sound data that can be exploited through alternative frameworks for supervised classification. First, bird sound is naturally decomposed into a collection of parts, e.g., syllables, which motivates a multi-instance learning (MIL) approach \cite{dietterich1997solving}. Second, multi-label classification (MLC) \cite{tsoumakas2007multi} is a natural fit for bird sound because an audio recording can be associated with a set of species (and other sounds) that are present. Multi-instance multi-label learning (MIML) combines both ideas. MIML has previously been used for classification of bird sound recordings containing multiple simultaneously vocalizing species \cite{briggs2012acoustic}. However, prior work on MIML for bird sound has focussed more on the multi-instance structure of the sound, and less on structure in the species/label sets.

The MLC framework has not been directly applied to bird sound (although some MIML algorithms which have been applied to bird sound can be considered a reduction to MLC, e.g., MIML-kNN \cite{zhang2010k} and MIML-RBF \cite{zhang2009mimlrbf}). Ensemble of classifier chains (ECC) \cite{read2011classifier} is an algorithm for MLC which has recently been applied to species distribution modeling, where the goal is to predict the set of bird species present at a site from a feature vector describing physical and biological properties of the site. Yu et al.~\cite{yu_mlc_sdm} suggested that ECC achieves better performance in this domain than binary relevance because it can exploit correlations in the label sets. Considering this observation, we hypothesize that ECC can exploit the same structure while predicting sets of bird species from an acoustic feature vector instead of environmental covariates. 

We formulate the classification problem similarly to \cite{briggs2012acoustic}. The training data consists of audio recordings paired with a set of species that are present. The goal is to predict the set of species in a new recording which is not part of the training data. 

To apply MLC, it is necessary to represent each audio recording with a fixed-length feature vector. We apply a  2D time-frequency supervised segmentation algorithm similar to \cite{lawrence,briggs2012acoustic}, then compute the same features as in \cite{briggs2012acoustic} to describe each segment. Then we use a clustered codebook to obtain a histogram-of-segments for each recording. \cite{somervuo2004bird} used histograms to represent variable-length sequences of syllables. \cite{briggs2009audio} used histograms of frame-level features (spectrum and MFCC) to represent an audio recording with a single species of bird.

We compare ECC, binary relevance (BR), and results from prior work on two real-world datasets of birdsong with multiple simultaneously vocalizing species.

The first dataset was collected with unattended omnidirectional microphones in the H.~J.~A.~(HJA) Experimental Research Forest, and has previously been used in several classification experiments \cite{briggs2012acoustic,briggs2012rank,briggs_tkdd,liu2012conditional} 

The second dataset is new, and consists of recordings of birds made with an iPhone in a residential neighborhood (collected and labeled by the authors). The new iPhone birdsong dataset presents the same multi-species issues as the HJA Birdsong dataset, but is arguably more challenging because there are more/louder sources of background noise and non-bird classes (especially motor vehicles and insects).

Results are analyzed in terms of standard multi-label error measures: Hamming loss, set 0/1 loss, rank loss, 1-error, and coverage. ECC achieves better results than BR in the majority of comparisons, and ECC with no parameter tuning is better than one and worse than two of the MIML algorithms (which have an unfair advantage of using post-hoc parameter tuning).

\section{Problem Statement}

In MLC, the training dataset is $(\mathbf{x}_1, Y_1), \ldots, (\mathbf{x}_n, Y_n)$
where $\mathbf{x}_i \in \mathbb{R}^d$ is a feature vector, and $Y_i \subseteq \mathcal{Y} =  \{1,\ldots, c\}$ is a subset of $c$ possible class labels.
The goal is to learn a classifier $f(\mathbf{x}) : \mathbb{R}^d \rightarrow 2^{\mathcal{\mathcal{Y}}}$ which predicts a label set from a given feature vector. It is common to implement and evaluate multi-label classifiers based on a score function for each class $f_j(\mathbf{x}) : \mathbb{R}^d \rightarrow \mathbb{R}$, which represents the predicted confidence that label $j$ is in the set. The set predictor $f$ is defined in terms of the score functions $f_1, \ldots, f_c$. The MLC framework maps to acoustic species classification as follows: each audio recording is associated with a feature vector, and the set of species audible in the recording is the label set.

MIML is a related framework where the training data consists of bags-of-instances paired with label sets,
\begin{eqnarray}
(B_1, Y_1), \ldots, (B_n, Y_n) \textrm{   where    } B_i = \{ \mathbf{x}_{i1}, \ldots, \mathbf{x}_{i n_i} \}
\end{eqnarray}
We will use MIML as an intermediate representation of audio recordings of bird sound, and solve the problem by a reduction from MIML to MLC.

\section{Background}

Binary relevance is one of the simplest algorithms for MLC. It is a reduction to SISL where binary prediction of each label is treated as a completely separate/independent problem. To refer to a bit in the binary representation of a label set, let $Y_i^j = I[j \in Y_i]$. BR creates $c$ SISL datasets $D_1, \ldots, D_c$, where $D_j = \{ (\mathbf{x}_i, Y_i^j) \}_{i=1}^n$, and trains a binary SISL classifier $f_j : \mathbb{R}^d \rightarrow \mathbb{R}$ on each $D_j$.

Classifier chains are also a reduction to SISL, but the problems for each class are not totally separate. CC predicts bits of the label set one at a time in a particular order, and uses all of the previously predicted bits as features for the next bit. CC creates $c$ SISL datasets $D_1, \ldots, D_c$, where
\begin{eqnarray}
D_j = \{ (\mathbf{x}_i \oplus Y_i^{1:j-1}, Y_i^j \}_{i=1}^n
\end{eqnarray} 
The notation $Y_i^{1:j-1}$ denotes the first $j-1$ bits of the binary representation of $Y_i$, and $\oplus$ is vector concatenation.
CC trains a binary SISL classifier $f_j : \mathbb{R}^{d + j - 1} \rightarrow \mathbb{R}$ on each dataset $D_j$. Algorithm \ref{alg:cc_classify} is pseudocode for classification of a feature vector $\mathbf{x}$ with CC. Assuming the SISL classifier $f_j$ outputs a score or probability, a threshold $t$ is used to make a 0/1 prediction.

\begin{algorithm}[tb]
   \caption{Classifier Chains -- classify $\mathbf{x}$}
   \label{alg:cc_classify}
\begin{algorithmic}
   \STATE $Y = []$
   \FOR{$j=1$ {\bfseries to} $c$}
   \STATE $Y = Y \oplus I[f_j(\mathbf{x}_i \oplus Y) > t]$
   \ENDFOR
\STATE {\bfseries return} $Y$
\end{algorithmic}
\end{algorithm}

ECC creates an ensemble of $L$ classifier chains, where each chain $l=1,\ldots,L$ views the classes in a different random permutation $\pi_l : \{1,\ldots,c\} \rightarrow \{1,\ldots,c\}$. Each chain in the ensemble votes on each potential class in the label set.
For each chain $l$ and class $j$, ECC trains a SISL classifier $f_{lj}$ on the dataset
\begin{eqnarray}
D_{lj} = \{ (\mathbf{x}_i \oplus Y_i^{\pi_l(1)} \oplus \ldots \oplus Y_i^{\pi_l(j-1)}, Y_i^{\pi_l(j)} \}_{i=1}^n
\end{eqnarray}

\section{Proposed Methods}

\subsection{Classifier Chains with Random Forest}

\begin{algorithm}[tb]
   \caption{ECC-RF -- class scores for $\mathbf{x}$}
   \label{alg:eccrf_classify}
\begin{algorithmic}

	\STATE $score[1,\ldots,c] = 0$
	
	\FOR{$l=1$ {\bfseries to} $L$}
	
	\STATE $\mathbf{x}' = \mathbf{x}$

	\FOR{$j=1$ {\bfseries to} $c$}
	   
	\STATE $p_{lj} = f_{lj}(\mathbf{x}')$
	\STATE $score[\pi_l(j)] = score[\pi_l(j)] + p_{lj}$
	   
	\IF{$j \neq 1$} 
		\STATE $\mathbf{x}' = \mathbf{x}' \oplus p_{lj}$
	\ENDIF
	   
	\ENDFOR
	   
	\ENDFOR
   
\STATE {\bfseries return} $scores / L$
\end{algorithmic}
\end{algorithm}

We implement ECC with a Random Forest (RF) as the base-SISL classifier, hence we call the proposed classifier ECC-RF. Because RF outputs a probability, the ensemble can be viewed as an instance of the Ensemble of Probabilistic Classifier Chains (EPCC) algorithm \cite{dembczynski2010bayes}. Therefore it is reasonable to aggregate probabilities from each SISL classifier rather than 0/1 votes. The aggregated probabilities are used as the score-functions for each class. Algorithm \ref{alg:eccrf_classify} gives pseudocode we use to generate a class-score vector with ECC-RF, given input $\mathbf{x}$. 

\subsection{Out-Of-Bag Calibrated Thresholds}

Sometimes class scores are sufficient, for example to rank species from most likely to least likely to be present. However, it is often desirable to obtain a specific predicted label set. A label set can be obtained by comparing each score to a threshold. The simplest method is to use a single threshold for all classes \cite{tsoumakas2007multi}. We instead select a separate threshold for each class, which is calibrated using out-of-bag (OOB) estimation \cite{breiman2001random} (for both BR and ECC-RF). Consider one of the binary RF's in BR or ECC-RF, $f_j$ or $f_{lj}$. Let its OOB estimate on instance $\mathbf{x}_i$ in the training dataset be $\hat{f}_{j}(\mathbf{x}_i, i)$ (for BR) or $\hat{f}_{lj}(\mathbf{x}_i, i)$ (for ECC). For each class $j$, we select a threshold $t_j$ to minimize the 0/1 error on that class, comparing ground-truth labels for class $j$ with OOB estimates. The threshold used in BR for class $j$ is
\begin{eqnarray}
t_j = \mathop{\arg \min}_{t \in \{.001, \ldots, .999\}}   \sum_{i=1}^n I[ I[\hat{f}_j(\mathbf{x}_i, i) > t] = Y_i^{j}]
\end{eqnarray}
The same algorithm is applied to ECC-BR by defining $\hat{f}_j = L^{-1} \sum_{l=1}^L \hat{f}_{jl}$.

\section{Experiments}

\subsection{Datasets}

Two real-world birdsong datasets are used in our experiments. 



\paragraph{HJA Birdsong} 
The HJA Birdsong dataset consists of 548 ten-second audio recordings collected in the H. J. A.~Experimental Research Forest, using Songmeter SM1 recording devices. There are 13 species in this dataset, with between 1 and 5 species per recording (2.144 average). The most common sources of noise in this dataset include streams and wind. Further details of this dataset are available in \cite{briggs2012acoustic}. \cite{briggs2012acoustic} used 5-fold cross-validation for this dataset. We use the same 5-fold partitions, so the results are comparable.

\paragraph{iPhone Birdsong} 
We collected 150 five-second audio recordings of bird sound with an iPhone 4G in a residential neighborhood. 54 of the recordings were collected during the dawn chorus on a single day, and the rest were collected at different times of day over several months in 2012--13.

We filtered the original 150 recordings down to 91 which are more suited for a cross-validated species classification experiment.
There were 32 recordings with bird species we were unable to identify, and many more with non-bird sounds.
We removed all recordings containing unknown bird species, amphibians, human voice, dogs barking, and the iPhone vibrating due to receiving a message. Finally, we remove all recordings containing a species which appears only once in the dataset (cross-validation is not reasonable in this case). The filtered subset of 91 recordings contains 14 species. Many of these recordings still contain motor vehicle noise, loud insects, and ``click noises'' which appear as vertical lines in the spectrogram.
Table \ref{tbl:iphone_dataset} lists each species, and the number of recordings it appears in. Note that the dataset is highly unbalanced.
Because this is smaller dataset, we use 10-fold cross-validation instead of 5-fold.

\begin{table}[t]
\caption{The number of recordings containing each species in the iPhone Birdsong dataset.}
\label{tbl:iphone_dataset}
\begin{center}
\begin{small}
\begin{tabular}{lc}
Species & Recordings \\
\hline
American Goldfinch&2			\\
American Robin&23				\\
Black Capped Chickadee&36		\\
Black-headed Grosbeak&2			\\
Chestnut Backed Chickadee&3		\\
Golden Crowned Kinglet&6		\\
Great Horned Owl&2				\\
Killdeer&7						\\
Marsh Wren&3					\\
Northern Flicker&4				\\
Red Breasted Nuthatch&19		\\
Red-Winged Blackbird&23			\\
Spotted Towhee&13				\\
Stellar's Jay&4					\\
\end{tabular}
\end{small}
\end{center}

\vspace{-20pt} 

\end{table}

\subsection{Histogram-of-Segments Representation}


In order to apply MLC, we represent each audio file with a fixed-length feature vector. Prior work \cite{briggs2012acoustic} has shown that 2D time-frequency segmentation of a spectrogram is useful for separating bird sounds which may overlap in time. For the new iPhone Birdsong dataset, we follow a  similar process to \cite{briggs2012acoustic} for supervised 2D segmentation of spectrograms.\footnote{There are some minor differences in segmentation in the iPhone dataset vs.~the HJA dataset. For the iPhone dataset, the RF used for segmentation was trained on features consisting of pixels in an 17x17 window, the $y$-coordinate of the window center, and the average intensity in the window. This RF used 100 trees with a maximum depth of 10. We annotated 20 out of 91 of the spectrograms in the dataset with examples of correct segmentation.} 

Each segment is isolated, and described by the same 38 acoustic features as in \cite{briggs2012acoustic}. At this point, the audio dataset is represented as a MIML dataset (each recording is a bag of segments paired with a set of species). We reduce this MIML dataset to an MLC dataset by summarizing all of the segments in a recording with a histogram. Hence, the feature vector used for MLC has dimension $k$, where $k$ is the number of clusters. For the HJA Birdsong dataset, we use the original segmentation and segment features from \cite{briggs2012acoustic}, rather than our slightly modified segmentation.

\begin{figure}[ht]
\begin{center}
\centerline{\includegraphics[width=\columnwidth]{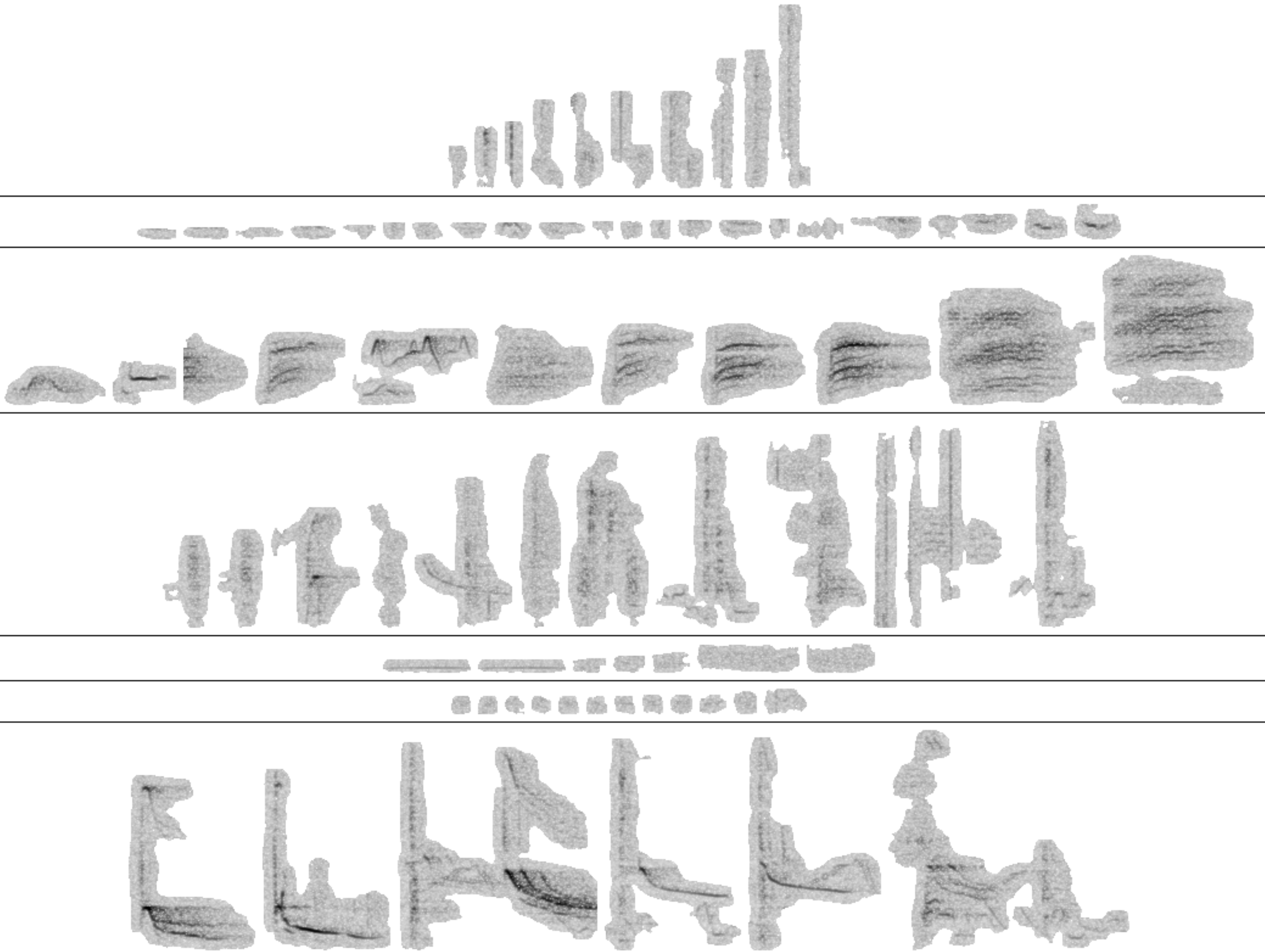}}
\caption{Example clusters of segments in the codebook used in the construction of histogram-of-segment features for the iPhone dataset (modified to enhance contrast).}
\label{cluster_examples}

\vspace{-12pt}

\end{center}
\end{figure} 

Segment features are clustered using $k$-means++ \cite{arthur2007k} to form a codebook. For each recording, each of its segments is mapped to a cluster center, and the normalized count of segments for each cluster is used as the histogram-of-segments feature. Figure \ref{cluster_examples} shows some example clusters from the codebook for the iPhone dataset.

\subsection{Comparison to MIML}

Using results from \cite{briggs2012acoustic} on the HJA dataset, we compare our proposed ECC-RF algorithm to three MIML algorithms: MIMLSVM, MIML-$k$NN and MIMLRBF. Each of these algorithms are reductions from MIML to MLC; they construct a single fixed-length feature vector from a bag of instances (i.e., a recording containing a varying number of segments), then apply binary relevance. For BR, MIMLSVM uses SVM as the base-SISL classifier, while MIML-$k$NN and MIMLRBF use linear models trained by unregularized min-squared-error. These MIML classifiers focus mainly on construction of a good ``summary'' feature vector, while using only the simplest MLC classifier. In contrast, our proposed method uses a simpler feature vector construction, and a more complicated model of structure in the label sets. 

\subsection{Parameters}

For constructing histogram of segment features, the parameter to $k$-means++ is $k=50$. 

The only parameters for ECC-RF are $L$, the number of chains, and $T$, the number of trees in each RF. It is expected that as these parameters are increased, the accuracy of the classifier converges to some asymptotic value. Hence selection of these parameters is mainly a matter of how much computation time is available. We conservatively chose $L=25, T=25$, and did no further optimization of these parameters.\footnote{Running 10 repetitions of 5- or 10-fold CV on both datasets with BR and ECC-RF takes 424 seconds on a Mac Pro with 2x2.4 GHz Quad-Core Intel Xeon Processors. The RF tree induction is parallel and the rest is sequential. The implementation is in C++ compiled with GCC 4.2.}

For BR, the only parameter is $T$, the number of trees in each RF. We set $T=25^2$ for BR to ensure that the total number of trees which cast a vote in every prediction is the same between BR and ECC-RF. All decision trees used in both BR and ECC use a maximum tree-depth of 15, and store histograms of class labels in decision tree leaves instead of the majority label.

The three MIML algorithms that we compare to in this experiment have parameters which must be tuned (e.g., by grid search). These tuning parameters are unlike the parameters of ECC-RF. Although such parameters can be optimized by cross-validation (with respect to a particular multi-label performance measure), doing so adds an order of magnitude runtime to the classification experiment, so \cite{briggs2012acoustic} used ``post-hoc'' parameter selection. In post-hoc selection, the experiment is run multiple times for all combinations of parameter values in a grid, and the best result from any parameter is reported. Therefore the MIML algorithms have an advantage in these experiments. 

\subsection{Results}

\begin{table*}[t]
\caption{Multi-label classification experiment results, averaged over 10 repetitions.\\ $\dagger$ -- Results from \cite{briggs2012acoustic} using post-hoc parameter selection.}
\label{tbl:results}
\begin{center}
\begin{small}
\begin{tabular}{ccccccc}
Dataset & Classifier 			& Hamming loss $\downarrow$ & Set 0/1 Loss  $\downarrow$	& Rank loss  $\downarrow$		& 1-error $\downarrow$ 			 & Coverage $\downarrow$ \\

\hline

iPhone Birdsong & BR-RF 	& 0.1148	 	&	0.811		&	0.1948		&	0.5154		&	3.5495 \\
iPhone Birdsong & ECC-RF 	& 0.1168 		&	0.8121	 	&	0.1927		&	0.5132 		&	3.5319 \\
	
\hline

HJA Birdsong & BR -RF		& 0.0489 		& 	0.4476 		&	0.0258 		&	0.044 		&	1.6805 \\
HJA Birdsong & ECC-RF		& 0.0485		&	0.4369		&	0.0246		&	0.0482 		&	1.6555\\
	
\hline

HJA Birdsong & MIMLSVM $\dagger$ 	& 0.054 		& 	N/A			& 	0.033		& 	0.067		& 	1.844 \\
HJA Birdsong & MIMLRBF $\dagger$ 	& 0.049 		& 	N/A 			& 	0.022		& 	0.034		& 	1.632 \\
HJA Birdsong & MIML-$k$NN $\dagger$ 	& 0.039 		& 	N/A 			& 	0.019		& 	0.036		& 	1.589 \\

\end{tabular}
\end{small}
\end{center}
\end{table*}

Table \ref{tbl:results} lists results. Because RF and ECC are randomized, we run 10 trials, and report results averaged over all trials and folds of cross-validation.

Following recommendations in \cite{demvsar2006statistical}, we summarize results for multiple classifiers on multiple datasets by win-loss counts (and do not discard any result as ``insignificant''). However, unlike the scenario considered by \cite{demvsar2006statistical}, we compare MLC classifiers rather than SISL classifiers, so there are multiple performance measures. Because there are only a few datasets and more performance measures, we aggregate win/loss counts over all measures.

Comparing BR and ECC-RF on two datasets with five different performance measures gives 10 comparisons between the two algorithms. Over both datasets, the win-loss count for ECC-RF vs.~BR is 7-3.
On the iPhone dataset, the result is less decisive; the count for ECC-RF vs.~BR is 3-2. On the HJA Birdsong dataset, the count for ECC-RF vs.~BR is 4-1. Overall these results suggest there is an advantage to using ECC-RF over BR for multi-label classification of bird species sets, given the histogram-of-segments representation.

Next we consider the win-loss counts on the HJA Birdsong dataset for ECC-RF vs.~MIMLSVM, MIMLRBF, and MIML-$k$NN. The counts are 5-0, 1-4, and 0-5, respectively, i.e.~MIMLSVM is worse than ECC-RF in all comparisons, but MIMLRBF and MIML-$k$NN are better than ECC-RF. However, this is not an entirely fair comparison due to post-hoc parameter selection in the MIML experiments.

\section{Discussion}

We suggest that the performance advantage of MIMLRBF and MIML-$k$NN over ECC-RF may be attributed to better representation of the multi-instance structure in the data (compared to our histogram-of-segments representation). Based on comparisons between ECC and BR, better modeling of structure in the label set is beneficial when compared with the same features and base-SISL classifier. 

\section{Related Work}

We focussed on learning to predict species label sets. Another interesting problem is to train on recordings with multiple labels, but classify segments with a single label. Such an approach reduces the labeling effort required to train SISL segment/syllable classifiers such as \cite{fagerlund2007bird,damoulas2010bayesian}. This problem is naturally formulated in the framework of MIML instance annotation \cite{briggs2012rank,briggs_tkdd}. A related formulation is to associate each segment with a set of candidate labels, only one of which is correct. This formulation is called ambiguous label classification \cite{cour2011learning}, or superset label learning \cite{liu2012conditional}.



\bibliographystyle{icml2013}
\end{document}